\documentclass[runningheads]{llncs}

\usepackage{eccv}

\usepackage{eccvabbrv}

\usepackage{graphicx}
\graphicspath{{../}}
\usepackage{booktabs}
\usepackage{float}
\usepackage{placeins}
\usepackage[accsupp]{axessibility}  
\usepackage{verbatim}
\usepackage{multirow}

\usepackage{hyperref}

\usepackage{orcidlink}

\usepackage[table,dvipsnames]{xcolor}

\makeatletter
\newcommand{\equalcontrib}{\textsuperscript{\@fnsymbol{1}}}
\newcommand{\equaladvising}{\textsuperscript{\dag}}
\makeatother

\begin{document}

\title{SK-Adapter: Skeleton-Based Structural Control for Native 3D Generation} 

\titlerunning{SK-Adapter}

\author{Anbang Wang\inst{1}\thanks{Equal contribution. \textsuperscript{\dag} Equal advising.}\orcidlink{0009-0008-2527-707X} \and
Yuzhuo Ao\inst{1}\equalcontrib\orcidlink{0009-0006-2598-4511} \and
Shangzhe Wu\inst{2}\equaladvising\orcidlink{0000-0003-1011-5963} \and
Chi-Keung Tang\inst{1}\equaladvising\orcidlink{0000-0001-7155-2919}}

\authorrunning{A.~Wang et al.}

\institute{The Hong Kong University of Science and Technology\\
\and
University of Cambridge\\
\email{\{awangas,yaoaa\}@connect.ust.hk, sw2181@cam.ac.uk,
cktang@cs.ust.hk}
}

\maketitle
\begin{figure}
    \centering
\includegraphics[width=1.0\linewidth]{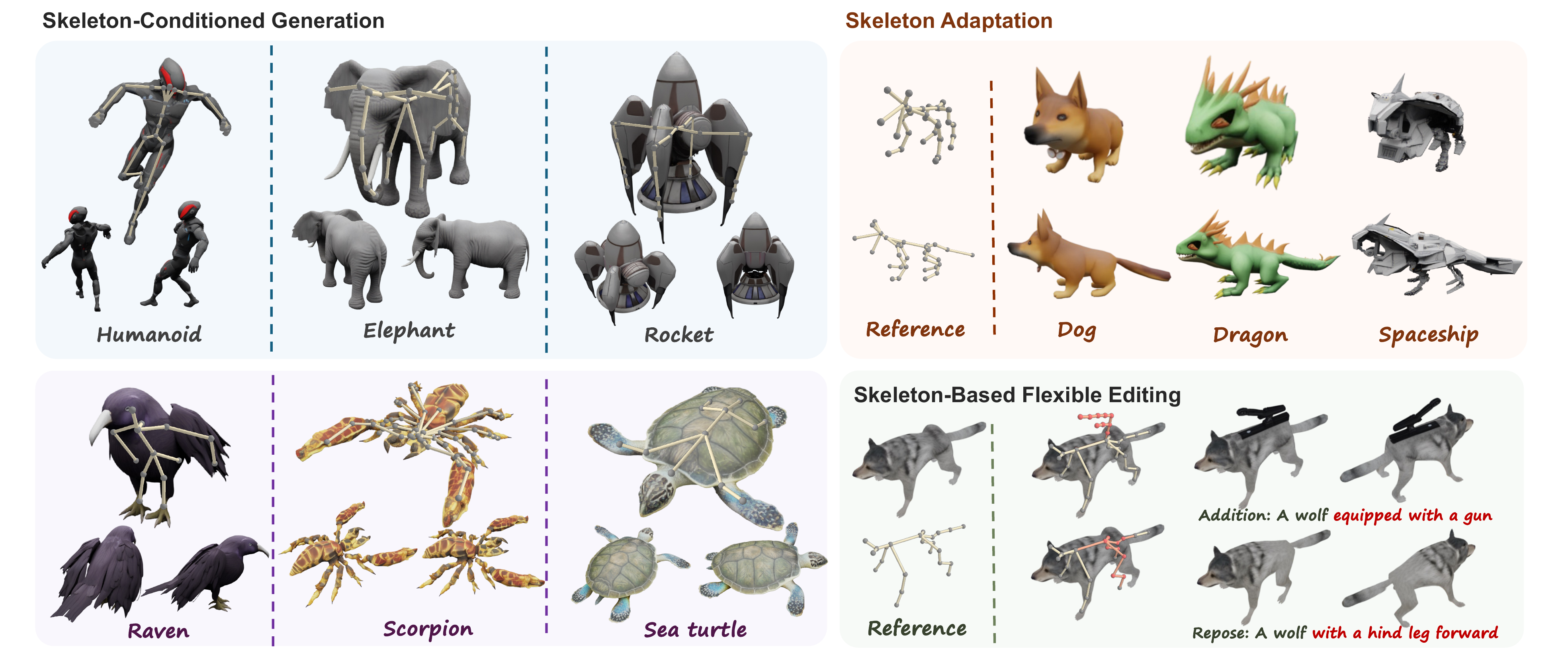} 
    \caption{SK-Adapter efficiently generates 3D assets in native 3D domain from given skeletons with its lightweight and effective designs, which also supports skeleton adaptation and flexible skeleton-based editing.}
    \label{fig:teaser}
\end{figure}

\begin{abstract}

      Native 3D generative models have achieved remarkable fidelity and speed, yet they suffer from a critical limitation:  inability to prescribe precise structural articulations, where precise structural control within the native 3D space remains underexplored. This paper proposes SK-Adapter, a simple yet efficient and effective framework that unlocks precise skeletal manipulation for native 3D generation. Moving beyond text or image prompts, which can be ambiguous for precise structure, we treat the 3D skeleton as a first-class control signal. SK-Adapter is a lightweight structural adapter network that encodes joint coordinates and topology into learnable tokens, which are injected into the frozen 3D generation backbone via cross-attention. This design allows the model to not only effectively ``attend'' to specific 3D structural constraints but also preserve its original generative priors. To bridge the data gap, we contribute the Objaverse-TMS dataset, a large-scale dataset of 24k text-mesh-skeleton pairs. Extensive experiments confirm that our method achieves robust structural control while preserving the geometry and texture quality of the foundation model, significantly outperforming existing baselines. Furthermore, we extend this capability to local 3D editing, enabling region-specific editing of existing assets with skeletal guidance, which is unattainable by previous methods. Project page: \url{https://sk-adapter.github.io/}
  
\end{abstract}

\section{Introduction}
\label{sec:intro}

The field of 3D content creation is undergoing a paradigm shift, transitioning from time-consuming optimization-based methods \cite{poole2022dreamfusiontextto3dusing2d, wang2023prolificdreamerhighfidelitydiversetextto3d} to efficient, feed-forward native 3D generative models \cite{hong2024lrmlargereconstructionmodel, lai2025hunyuan3d25highfidelity3d, xiang2025nativecompactstructuredlatents, xu2024instantmeshefficient3dmesh, xiang2025structured3dlatentsscalable,zhang2024claycontrollablelargescalegenerative}. These emerging frameworks, typically built upon scalable flow transformers, can synthesize high-fidelity 3D assets from text or images in mere seconds. However, despite their impressive visual quality and speed, precise structural controllability remains a formidable challenge. While text prompts convey high-level semantics and image prompts provide view-specific visual cues, both fall short in prescribing precise 3D articulations for entire assets, such as ``bending the knee 60 degrees'', or defining atypical anatomical topologies. For generated assets to be usable in animation and gaming pipelines, explicit structural control is indispensable.

Among various structural representations like bounding boxes or point clouds, skeletons are standard, compact, and hierarchical abstractions for articulated objects and are widely used for downstream tasks like animation. In the realm of 2D generation, the integration of skeletal guidance has achieved remarkable success. Pioneering works such as ControlNet\cite{zhang2023addingconditionalcontroltexttoimage} and T2I-Adapter\cite{mou2023t2iadapterlearningadaptersdig} demonstrated that injecting 2D human pose images into diffusion models allows for precise control over the layout and posture of generated images. Inspired by this success, recent 3D approaches like SKDream\cite{Xu_2025_CVPR} have attempted to replicate this paradigm to 3D generation by adopting a ``2D lifting'' strategy. They project arbitrary 3D skeletons into 2D projections to condition multi-view diffusion models, which are subsequently lifted to 3D via reconstruction.

However, directly transferring this 2D-conditioned paradigm to 3D generation introduces a fundamental dimensionality mismatch. While 2D skeletons work well for 2D images, compressing intrinsic 3D structures into 2D planes for 3D generation inevitably leads to spatial ambiguity—depth information is flattened, and self-occlusions in projected views cause the model to misinterpret complex topologies. Furthermore, relying on multi-stage reconstruction pipelines often degrades texture quality and introduces geometric artifacts, limiting the fidelity of the final assets.

To achieve precise structural fidelity without compromising generation quality, we posit that the control signal must be congruent with the generation space. Skeletal information should be injected directly within the native 3D domain, bypassing the lossy 2D projection bottleneck. However, realizing this goal faces the difficulty of adapting large-scale 3D transformers to follow strict skeletal structure constraints without catastrophic forgetting of their generative priors.

In this work, we propose {\bf SK-Adapter}, a novel framework for efficient, skeleton-guided native 3D generation. 
Our key insight is that precise structural control stems from the strict spatial alignment between the guidance signal and the generative latents. Instead of forcing the model to interpret abstract features or foreign 2D projections, we conceptualize the 3D skeleton as a set of sparse spatial tokens. These tokens encapsulate both geometric coordinates and topological constraints, which are then seamlessly injected into the transformer backbone through novel skeletal cross-attention layers, allowing the model to attend to precise 3D spatial constraints during volumetric generation.

Adapter methods, which originated in natural language processing and are now being quickly adopted for customizing pretrained transformer models by inserting lightweight modules, have 
yielded successful results in ControlNet~\cite{zhang2023addingconditionalcontroltexttoimage}, IP-Adapter~\cite{ye2023ipadaptertextcompatibleimage},
T2I-Adapter~\cite{mou2023t2iadapterlearningadaptersdig}, to name a few. Inspired by recent  studies\cite{ye2023ipadaptertextcompatibleimage, mou2023t2iadapterlearningadaptersdig, huang2024mvadaptermultiviewconsistentimage}, we employ this Parameter-Efficient Fine-Tuning (PEFT) strategy by freezing the parameters of the pre-trained backbone from Trellis\cite{xiang2025structured3dlatentsscalable} and training only the lightweight adapter modules, namely, the {\em skeleton encoder} and the {\em injected cross-attention layers}. This SK-Adapter design ensures that the model acquires a robust interpretation of skeletal structure, while fully preserving the powerful generative capabilities of the original foundation model. 

Beyond generation, SK-Adapter benefits from our disentangled structural control and has a high potential for flexible 3D editing. With our adapter, our method enables {\em tuning-free} operations such as addition and replacement of local regions, guided by skeleton prompts.
Fig.~\ref{fig:teaser} shows examples generated or edited by SK-Adapter to demonstrate this high potential.

In summary, our main contributions are:
\begin{enumerate}
    \item  We propose {\bf SK-Adapter}, the first framework that achieves skeletal control during native 3D generation. Extensive experiments demonstrate that our method significantly outperforms existing baselines in both structural alignment and generation quality.
    \item  We construct the {\bf Objaverse-TMS} dataset, comprising 24k high-quality text-mesh-skeleton triplets, addressing the data scarcity bottleneck in structure-guided 3D generation.
    \item We demonstrate the capability for flexible {\bf 3D editing}. SK-Adapter allows for precise region-specific local editing {\em based on skeleton prompts}.
\end{enumerate}

\section{Related Work}
\label{sec:related}

\noindent {\bf 3D generative models.} 
Recent breakthroughs in diffusion models \cite{ho2020denoisingdiffusionprobabilisticmodels,song2022denoisingdiffusionimplicitmodels,lipman2023flowmatchinggenerativemodeling}, and the increasing availability of large-scale, high-quality 3D datasets \cite{deitke2022objaverseuniverseannotated3d,deitke2023objaversexluniverse10m3d} have significantly accelerated the evolution of 3D generation. The field has shifted from time-consuming, per-asset optimization \cite{poole2022dreamfusiontextto3dusing2d,huang2023dreamwaltzmakescenecomplex} and multi-view reconstruction-based synthesis \cite{huang2026stereogsmultiviewstereovision, huang2024mvadaptermultiviewconsistentimage,liu2024syncdreamergeneratingmultiviewconsistentimages,long2023wonder3dsingleimage3d,tang2024lgmlargemultiviewgaussian,voleti2024sv3dnovelmultiviewsynthesis,wang2024crmsingleimage3d,xu2024instantmeshefficient3dmesh} toward native 3D generative models\cite{hong2024lrmlargereconstructionmodel, lai2025hunyuan3d25highfidelity3d, xiang2025nativecompactstructuredlatents, xu2024instantmeshefficient3dmesh, zhang2024claycontrollablelargescalegenerative,li2025craftsman3dhighfidelitymeshgeneration,li2025triposghighfidelity3dshape,wu2024direct3dscalableimageto3dgeneration,wu2025direct3ds2gigascale3dgeneration,zhao2023michelangeloconditional3dshape,zhang20233dshape2vecset3dshaperepresentation} like Trellis\cite{xiang2025structured3dlatentsscalable}. These models typically comprise a variational autoencoder (VAE) \cite{kingma2022autoencodingvariationalbayes} and a Diffusion Transformer (DiT) \cite{peebles2023scalablediffusionmodelstransformers} for denoising in latent space and directly operate on 3D structured latents or volumetric representations. These methods unify 3D generation with high fidelity and consistency, eliminating inconsistent multi-view synthesis or inefficient optimization.

\noindent {\bf Derivatives and extensions of 3D generative models.}
Following the success of foundational 3D generative frameworks, numerous derivatives have emerged to enhance structural granularity, applicability and editability. To improve generation quality and representational efficiency, works\cite{lai2025latticedemocratizehighfidelity3d,jin2025uniartunified3drepresentation,jia2025ultrashape10highfidelity3d} like Ultra3D\cite{chen2025ultra3defficienthighfidelity3d} have explored advanced architectural designs and diverse 3D priors. To achieve finer control, part-aware methods\cite{tang2025efficientpartlevel3dobject,yang2025holopartgenerative3damodal,chen2025autopartgenautogressive3dgeneration,dong2025morecontextuallatents3d,yan2025xparthighfidelitystructure} such as OmniPart\cite{yang2025omnipartpartaware3dgeneration}, and BANG\cite{Zhang_2025} introduce generative mechanisms for part-level decomposition. Meanwhile, the integration of 3D reconstruction with generation, as seen in Amodal3R\cite{wu2025amodal3ramodal3dreconstruction}, SAM3D\cite{sam3dteam2025sam3d3dfyimages} facilitates more robust reconstruction from partial observations. In terms of manipulation, zero-shot editing frameworks\cite{xiang2025structured3dlatentsscalable} such as VoxHammer\cite{li2025voxhammertrainingfreeprecisecoherent} and Nano3D\cite{ye2025nano3dtrainingfreeapproachefficient} enable high-fidelity attribute modifications without extensive retraining.

\noindent {\bf Skeleton-guided generation.}
Early work AnimatableDreamer \cite{wang2024animatabledreamertextguidednonrigid3d} generates 3D objects with unstructured skeletons extracted from given videos by canonical score distillation. Many works\cite{ye2023ipadaptertextcompatibleimage,ju2023humansdnativeskeletonguideddiffusion,huang2024dreamwaltzgexpressive3dgaussian,wang2024discodisentangledcontrolrealistic,hu2024animateanyoneconsistentcontrollable,tan2024animatexuniversalcharacterimage,wang2025poseanythinguniversalposeguidedvideo,zhang2023addingconditionalcontroltexttoimage,mou2023t2iadapterlearningadaptersdig } in 2D domain have demonstrated that injecting 2D human pose skeletons into diffusion models enables precise manipulation of layout and posture or consistent motion generation for image and video generation. Inspired by this success in 2D domain, a recent work SKDream\cite{Xu_2025_CVPR}  adopts a ``2D lifting'' strategy, where arbitrary 3D skeletons are projected into 2D maps to condition a multi-view diffusion model\cite{shi2024mvdreammultiviewdiffusion3d}, followed by 3D reconstruction\cite{xu2024instantmeshefficient3dmesh} and UV refinement to generate 3D assets conditioned on the given skeleton. However, it suffers from spatial ambiguity in 2D skeleton projection and inconsistency across multi-view images during reconstruction.  A concurrent work, Posemaster\cite{Yan_2026_CVPR}, focuses on humanoids and generates 3D humanoids conditioned on a fixed humanoid skeleton with diverse poses.

\section{Method}
\subsection{Problem Formulation}
The goal of skeleton-guided 3D generation is to synthesize a high-fidelity 3D asset $\mathcal{X}$ (represented as a 3D latent or mesh) that is consistent with both a textual prompt $\mathcal{T}$ and a specific structural guidance provided by a 3D skeleton $\mathcal{S}$. 

Formally, a 3D skeleton $\mathcal{S}$ is defined as a structured tuple $\mathcal{S} = \{J, G\}$, where:
\begin{itemize}
    \item $J \in \mathbb{R}^{N \times 3}$ denotes the spatial coordinates of $N$ joints in 3D space.
    \item $G$ represents the topology graph, typically defined by a parent mapping function $p(i)$ or an adjacency matrix $A \in \{0, 1\}^{N \times N}$, capturing the kinematic constraints.
\end{itemize}
The objective is to learn a conditional mapping $\mathcal{F}: (\mathcal{T}, \mathcal{S}) \to \mathcal{X}$. The primary challenge lies in preserving the intricate structural details prescribed by skeletal structure while maintaining the generative quality and diversity underlying the large-scale 3D priors. 

\label{sec:method}

\begin{figure}[t]
    \centering
    \includegraphics[width=1.0\linewidth]{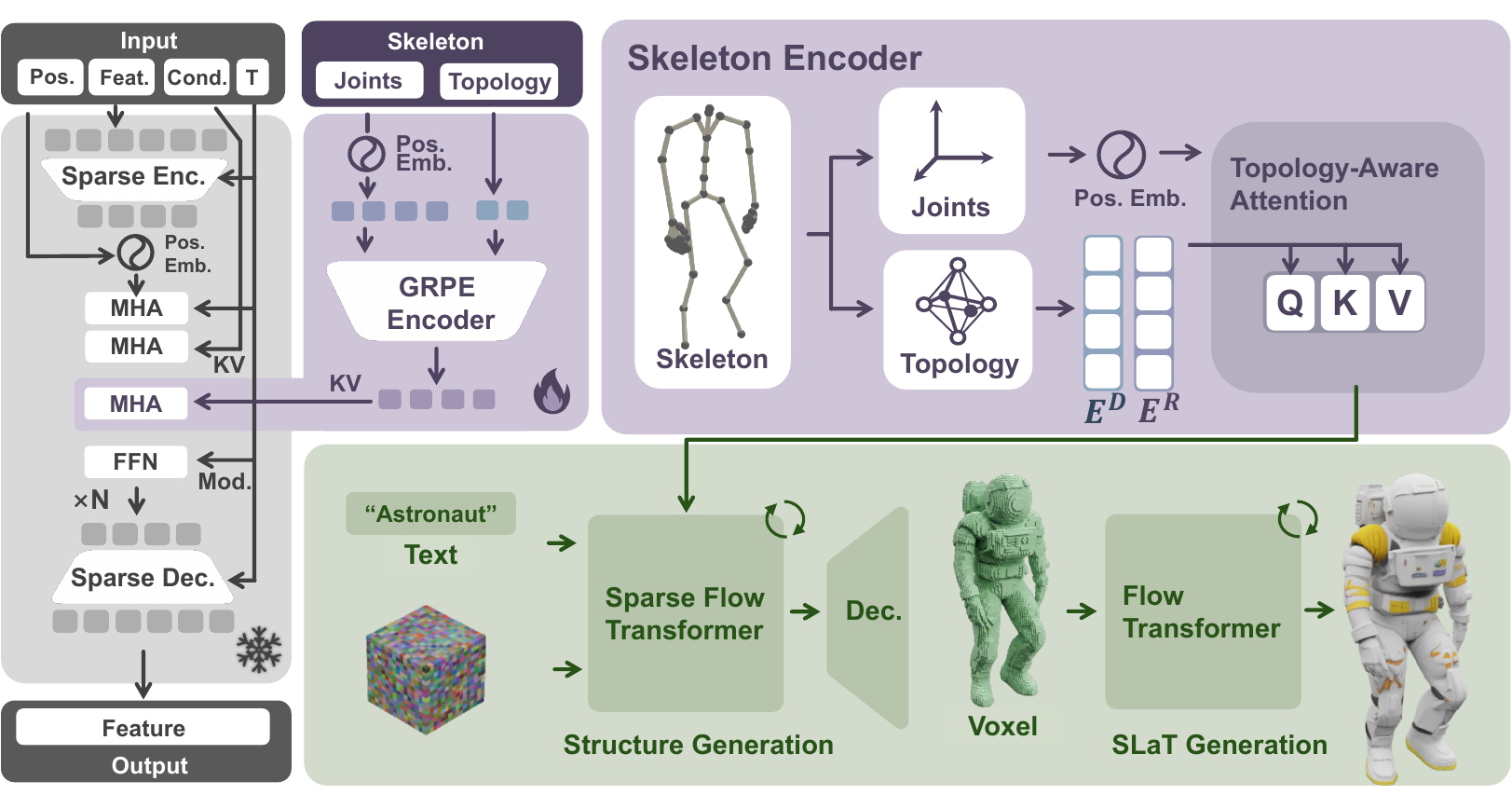} 
    \caption{Overview of the SK-Adapter framework. The GRPE module encodes the 3D skeleton's joints and topology into sparse tokens. These tokens are injected into a frozen pre-trained backbone via trainable cross-attention layers.}
    \label{fig:sk_adapter}
\end{figure}
\subsection{SK-Adapter}
We propose SK-Adapter, a 3D skeleton-guided generation framework for precise
structural control. As shown in Fig. \ref{fig:sk_adapter}, unlike heavy multi-stage pipelines that rely on ambiguous 2D
projections, SK-Adapter utilizes Trellis~\cite{xiang2025structured3dlatentsscalable} as backbone and injects joint-based positional tokens into
sparse structure transformer blocks. This simple yet effective approach
ensures spatial accuracy and high data efficiency, enabling the generation of
diverse, precisely controlled 3D assets in seconds.
\subsubsection{Topology-Aware Encoding.} 
We represent the input skeleton as a graph $\mathcal{G} = (\mathcal{V}, \mathcal{E})$, where $\mathcal{V}$ denotes the set of joint coordinates $J \in \mathbb{R}^{N \times 3}$ and $\mathcal{E}$ represents the hierarchical connectivity $G$. To capture the intricate spatial and structural relationships within the skeleton, we employ Graph Relative Positional Encoding (GRPE)~\cite{park2022grperelativepositionalencoding}. We integrate graph properties into the attention maps using two types of node affinity: the topological distance $\mathcal{D}_S$ and relations $\mathcal{R}_S$.

Specifically, for a given pair of joints $i, j \in J$, let $q_i$ and $k_j$ be the query and key representations. We learn distinct embeddings for topological distances $E_q^D, E_k^D \in \mathbb{R}^{(d_{\it{max}}+1) \times F}$ and for relations $E_q^R, E_k^R \in \mathbb{R}^{6 \times F}$, where $F$ is the latent feature size. These embeddings are used to form two structural attention biases:
\begin{equation}
    a_{ij}^{D} = q_i \cdot E_{q}^{D}[D_{ij}] + k_j \cdot E_{k}^{D}[D_{ij}],
\end{equation}
\begin{equation}
    a_{ij}^{R} = q_i \cdot E_{q}^{R}[R_{ij}] + k_j \cdot E_{k}^{R}[R_{ij}],
\end{equation}
where $D_{ij}$ and $R_{ij}$ denote the topological distance and relation between joints $i$ and $j$, respectively, and $[\cdot]$ denotes an index in the embedding matrix. These terms are aggregated into the standard attention map, and the final scaled attention score is defined as:
\begin{equation}
    a_{ij} = \frac{q_i \cdot k_j + a_{ij}^{D} + a_{ij}^{R}}{\sqrt{F}}.
\end{equation}

The node features are further enriched by encoding graph information into the values:
\begin{equation}
    z_i = \sum_{j=1}^{J} \hat{a}_{ij} (v_j + E_{v}^{D}[D_{ij}] + E_{v}^{R}[R_{ij}]),
\end{equation}
where $\hat{a}_{ij}$ is the normalized attention weight. This mechanism yields a skeletal embedding $\mathbf{f}_{\it{skel}} \in \mathbb{R}^{J \times F}$ that integrates geometric coordinates with hierarchical topology.

\subsubsection{Skeletal Cross Attention Mechanism.}
To bridge the gap between the 3D skeletal domain and the 3D voxel domain, we introduce a Cross-Attention mechanism. In each block of the flow transformer, the intermediate voxel feature map $\mathbf{h}_{\it{base}}$ from the pre-trained backbone is used to query the skeletal information. Consequently, with the standard attention operation:
\begin{equation}
    \mathbf{f}_{attn}=\text{Attention}(Q, K, V) = \text{softmax}\left(\frac{QK^T}{\sqrt{d_k}}\right)V,
\end{equation}
where the Query ($Q$) is derived from the voxel features $\mathbf{h}_{base}$ of the frozen backbone, while the Key ($K$) and Value ($V$) are derived from the encoded skeletal features $\mathbf{f}_{skel}$ produced by the GRPE encoder, it allows each spatial voxel to dynamically ``attend'' to the most relevant skeletal joints, effectively mapping the topological constraints onto the 3D coordinate grid. 

To maintain the generative fidelity of the pre-trained model and ensure training stability, the output of the cross-attention layer, $\mathbf{f}_{\it{attn}}$, is integrated into the backbone via a non-invasive strategy. We pass $\mathbf{f}_{\it{attn}}$ through a zero-initialized linear layer $\mathbf{W}_{o}$ to obtain $\mathbf{h}_{\it{attn}}$. The final hidden state $\mathbf{h}'$ is updated via a residual connection:
\begin{equation}
    \mathbf{h}' = \mathbf{h} + \mathbf{h}_{\it{attn}} = \mathbf{h} + \mathbf{W}_{o}\mathbf{f}_{\it{attn}}.
\end{equation}
This design ensures that at the onset of training, SK-Adapter contributes a ``null signal,'' allowing the model to preserve the high-quality generative priors of the original Trellis pipeline. As optimization progresses, the layer gradually learns to modulate the voxel latents according to the skeletal guidance without destabilizing the pre-trained distribution.

\subsubsection{Training.} The training of SK-Adapter follows the Latent Flow Matching (LFM)\cite{dao2023flowmatchinglatentspace} paradigm, supervised in the compact latent space of a pre-trained 3D Voxel Autoencoder. For a ground-truth 3D asset, we first obtain its sparse latent representation $\mathbf{z}_0$ using the frozen Voxel Encoder $\mathcal{E}$. We then define a probability path between Gaussian noise $\mathbf{z}_1$ and the target latent $\mathbf{z}_0$. The SK-Adapter-enhanced model $v_\theta$ is trained to predict the velocity field that transforms the noise toward the skeletal-conditioned target:
\begin{equation}
    \mathcal{L}_{FM} = \mathbb{E}_{t, \mathbf{z}_0, \mathbf{z}_t} \left\| v_\theta(\mathbf{z}_t, t, \mathbf{c}_{\it{text}}, \mathbf{f}_{\it{skel}}) - \mathbf{u}_t(\mathbf{z}_0) \right\|^2,
\end{equation}
where $\mathbf{u}_t(\mathbf{z}_0)$ is the target velocity. During training, the entire transformer backbone remains frozen. Only the GRPE encoder, cross-attention layers, and zero-initialized projection layers are optimized. This prevents catastrophic forgetting of the base model's 3D knowledge while enabling precise structural control.

\subsection{Editing}

The locality of SLaT  allows for region-specific editing by altering voxels and latents in masked areas while leaving other areas intact. To this end, following Trellis\cite{xiang2025structured3dlatentsscalable}, we adapt Repaint\cite{lugmayr2022repaintinpaintingusingdenoising} to our skeleton-conditioned editing. Unlike previous methods that solely rely on text or image prompts, we additionally utilize the skeleton prompt for better structural control. 

Given a modified skeleton (e.g., with altered joint angles or topology) and a masked bounding box, we modify the flow matching sampling process to regenerate content strictly within these areas. The generation is conditioned on the unchanged background and the updated skeletal tokens via our SK-Adapter. Consequently, the first stage updates the structural topology to match the new skeletal guidance, and the second stage produces coherent surface details, enabling tuning-free operations such as precise addition and re-posing (replacement of skeleton).

\section{Experiments}
\label{sec:experiments}

\subsection{Dataset}
\begin{figure}[t]
    \centering
    \includegraphics[width=0.9\linewidth]{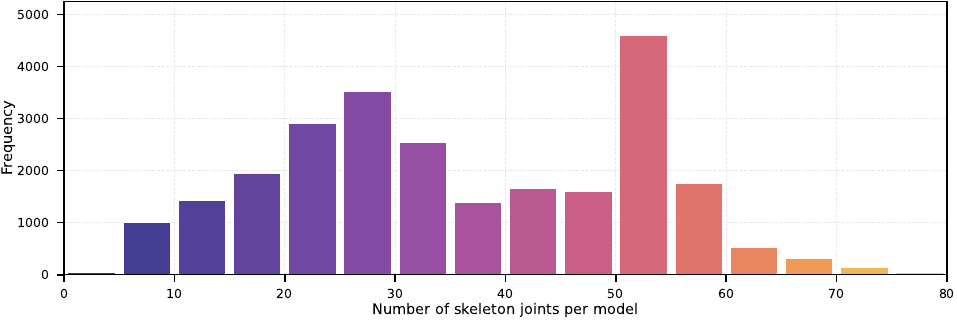}
    \caption{Distribution of skeleton joint counts in Objaverse-TMS.}
    \label{fig:dataset_statistics}
\end{figure}

\begin{figure}[t]
    \centering
    \includegraphics[width=\linewidth]{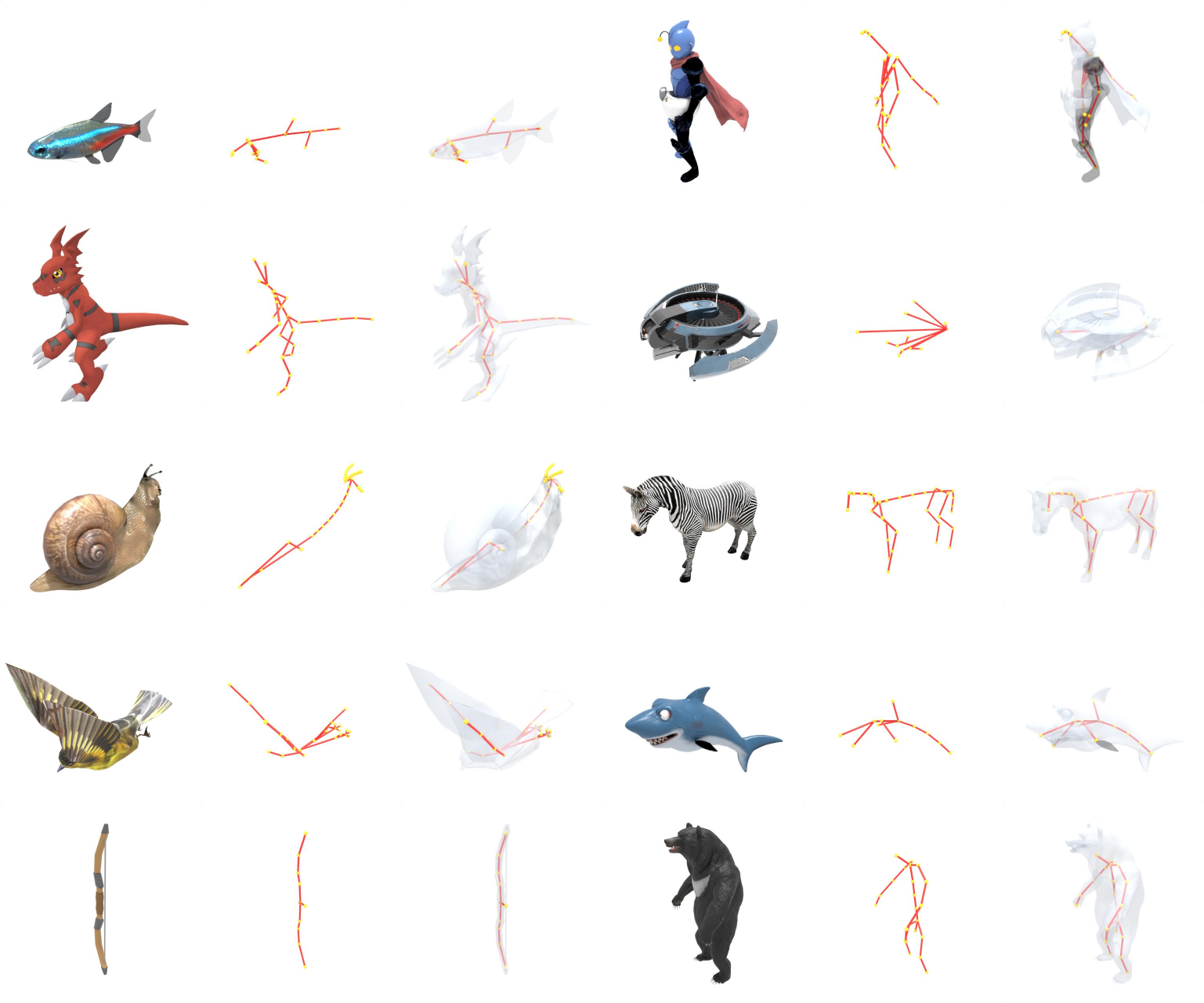}
    \caption{Representative samples from Objaverse-TMS. Each example shows the rendered 3D asset, its expert-annotated skeleton, and the skeleton-mesh overlay.}
    \label{fig:dataset_visualization}
\end{figure}

Training an effective model for skeleton-based 3D generation necessitates synchronized data across three modalities: textual descriptions, 3D meshes, and their corresponding skeletal rigs. However, the absence of such a tripartite dataset in the existing literature poses a significant bottleneck. While recent efforts like SKDream \cite{Xu_2025_CVPR} attempt to bridge this gap using autonomous skeleton generation with deterministic algorithms, these automatically generated rigs often suffer from anatomical inconsistencies and lack physical plausibility, which severely limits the model's ability to learn precise structural priors. To facilitate the training of our proposed framework, we curate Objaverse-TMS, a large-scale collection specifically designed for this task. We build Objaverse-TMS from Anymate\cite{deng2025anymatedatasetbaselineslearning} and CAP3D\cite{luo2023scalable3dcaptioningpretrained}, which are both based on the subsets of Objaverse-v1\cite{deitke2022objaverseuniverseannotated3d} and Objaverse-XL\cite{deitke2023objaversexluniverse10m3d}.  By extracting skeletal structures and captions from Anymate and CAP3D respectively, we establish a high-quality intersection of these modalities. We then process the original raw assets from Objaverse-v1 and Objaverse-XL to derive meshes and normalized voxels with skeletons. After filtering out samples with incomplete rigging, the resulting Objaverse-TMS dataset comprises 24K text-mesh-skeleton triplets, providing a robust foundation for skeleton-conditioned learning. We compare the existing datasets with skeleton annotations in Table \ref{tab:dataset_comparison}, report the skeleton joint-count statistics in Fig.~\ref{fig:dataset_statistics}, and visualize representative dataset samples in Fig.~\ref{fig:dataset_visualization}.

\begin{table}[t]
        \centering
        \caption{Comparison of Objaverse-TMS with other 3D datasets with skeleton annotations. }
        \label{tab:dataset_comparison}
        \begin{tabular}{lccc}
          \toprule
          \textbf{Dataset} & \textbf{Scale} & \textbf{Skeleton}  & \textbf{Text Caption} \\
          \midrule
          Rig-XL \cite{zhang2025modelrigalldiverse} & 14k & Expert & $\times$ \\
          Articulation-XL \cite{song2025magicarticulatemake3dmodels} & 33k & Expert & $\times$ \\
          Articulation-XL2.0 \cite{song2025puppeteerriganimate3d} & 59k & Expert & $\times$ \\
          Anymate \cite{deng2025anymatedatasetbaselineslearning} & 230k & Expert & $\times$ \\
          Objaverse-SK \cite{Xu_2025_CVPR} & 24k & Autonomous & \checkmark \\
          TextuRig \cite{zhao2026stroke3dlifting2dstrokes} & 7k & Expert & \checkmark \\
          \midrule
          \textbf{Objaverse-TMS} (Ours) & 24k & Expert & \checkmark \\
          \bottomrule
        \end{tabular}
        
\end{table}

As shown in Fig.~\ref{fig:dataset_visualization}, Objaverse-TMS covers a broad range of articulated 3D assets, including humanoids, animals, and other object categories with diverse geometric layouts. The paired skeletons remain spatially aligned with the corresponding meshes, providing explicit joint and bone structures that complement textual descriptions and enable the model to learn fine-grained skeletal control in native 3D space.

\subsection{Implementation Details}
 We train our SK-Adapter on Objaverse-TMS dataset for 200 epochs, with a batch size of 16 and a learning rate of $1\times10^{-5}$. To support classifier-free guidance, we apply a 10\% dropout to the text conditioning, while the skeleton conditioning remains dropout-free.

\subsection{Evaluation Protocol}
To ensure a rigorous and unbiased evaluation, we curate a diverse testing set comprising 140 test instances (TMS-eval) sampled from the validation split of Objaverse-TMS. To thoroughly assess the model's generalization capabilities across various topological complexities, this benchmark is strictly balanced to include three primary categories: humanoids, animals, and other objects. Specifically, these assets consist of 54 humanoid figures, 63 animals and 23 objects. 

\subsection{Evaluation Metrics} 
We evaluate our method and baselines in terms of structural/textual alignment and visual fidelity. All 2D-based metrics are computed from 12 uniformly distributed renderings of each generated 3D mesh.

\textbf{Structural and Textual Alignment.} To assess conditioning fidelity, we evaluate both structural and textual consistency. For structural alignment, we introduce the ReRigging Score: we apply the rigging model from Anymate~\cite{deng2025anymatedatasetbaselineslearning} to estimate a skeleton $\hat{S}$ from each generated mesh, then compute its Chamfer Distance to the conditioning skeleton $S$. Lower scores indicate better structural adherence; applying Anymate to ground-truth meshes gives an oracle reference of 0.2073. For textual alignment, we use CLIP Score to measure the semantic consistency between multi-view renders and text prompts.

\textbf{Visual Fidelity.} To assess the overall visual quality and realism of the generated geometries, we employ PickScore \cite{kirstain2023pickapic}  and Kernel Distance (KD-DINO). PickScore evaluates the human-aligned perceptual quality of the generated outputs. To evaluate distributional visual fidelity, we compute KD-DINO by extracting features from the rendered views using the DINOv2\cite{oquab2024dinov2learningrobustvisual} backbone, which robustly measures the global discrepancy between the generated multi-view image distribution and the ground-truth references.

\subsection{Baselines Comparison} 
We benchmark our SK-Adapter, a native but lightweight 3D approach, against representative, state-of-the-art structure-guided generative frameworks. Specifically, we compare against {SKDream}~\cite{Xu_2025_CVPR}, a multi-view diffusion approach that conditions generation on 2D projected maps of 3D skeletons prior to performing 3D reconstruction. We also modify another native 3D generation method,  {SpaceControl}\cite{fedele2025spacecontrolintroducingtesttimespatial} to our task for comparison. Specifically, SpaceControl is a training-free spatial guidance method built upon the Trellis\cite{xiang2025structured3dlatentsscalable}. SpaceControl injects structural guidance directly into the inference process. To adapt this method to our skeletal conditioning task, we convert the input sparse skeleton into a volumetric mesh representation by modeling joints as spheres and bones as cylinders with the same radius. The process starts from adding noise on conditional skeletons with a time step $t_s$ (set as $0.52$ to balance skeleton alignment and generation quality). Then clean 3D assets are generated by denoising steps.

\begin{table*}[t]
\centering

\caption{{Quantitative Comparison on TMS-eval (Structural and Textual Alignment).} We evaluate the methods based on Structural Alignment (ReRigging Score) and Textual Alignment (CLIP Score) across overall, humanoid, animal, and other categories. $\downarrow$ indicates lower is better, $\uparrow$ indicates higher is better. \textbf{Bold} indicates the best performance.}
\label{tab:alignment_results}
\resizebox{\textwidth}{!}{%
\begin{tabular}{l|cccc|cccc}
\toprule
 & \multicolumn{4}{c|}{\textbf{ReRigging Score} $\downarrow$} & \multicolumn{4}{c}{\textbf{CLIP Score} $\uparrow$} \\
\cmidrule{2-5} \cmidrule{6-9}
\textbf{Method} & Overall & Humanoid & Animal & Other & Overall & Humanoid & Animal & Other \\
\midrule
SKDream~\cite{Xu_2025_CVPR} & 0.2818 & 0.2385 & 0.2730 & 0.4075 & 25.65 & 25.33 & 26.34 & \textbf{24.52} \\
SpaceControl~\cite{fedele2025spacecontrolintroducingtesttimespatial} & 0.2740 & 0.2282 & 0.2898 & 0.3386 & 25.66 & 25.64 & 26.50 & 23.43 \\
\midrule
\textbf{SK-Adapter (Ours)} & \textbf{0.2228} & \textbf{0.1740} & \textbf{0.2415} & \textbf{0.2861} & \textbf{26.16} & \textbf{26.28} & \textbf{27.05} & 23.43 \\
\bottomrule
\end{tabular}
}

\caption{{Quantitative Comparison on TMS-eval (Visual Fidelity).} We evaluate the methods based on Visual Fidelity (PickScore, KD-DINO) across overall, humanoid, animal, and other categories. $\downarrow$ indicates lower is better, $\uparrow$ indicates higher is better. \textbf{Bold} indicates the best performance.}

\label{tab:visual_results}
\resizebox{\textwidth}{!}{%
\begin{tabular}{l|cccc|cccc}
\toprule
 & \multicolumn{4}{c|}{\textbf{PickScore} $\uparrow$} & \multicolumn{4}{c}{\textbf{KD-DINO} $\downarrow$} \\
\cmidrule(lr){2-5} \cmidrule(lr){6-9}
\textbf{Method} & Overall & Humanoid & Animal & Other & Overall & Humanoid & Animal & Other \\
\midrule
SKDream~\cite{Xu_2025_CVPR} & 20.46 & 20.01 & 20.88 & 20.40 & 1.3809 & 2.1106 & 1.7809 & 3.2435 \\
SpaceControl~\cite{fedele2025spacecontrolintroducingtesttimespatial} & 20.55 & 20.18 & 20.90 & \textbf{20.45} & 1.7821 & 2.1719 & 2.5034 & 3.6304 \\
\midrule
\textbf{SK-Adapter (Ours)} & \textbf{21.01} & \textbf{20.74} & \textbf{21.45} & 20.44 & \textbf{0.7778} & \textbf{1.1998} & \textbf{1.2673} & \textbf{2.3765} \\
\bottomrule
\end{tabular}
}
\end{table*}
\begin{figure}[tbp]
    \centering
    \includegraphics[width=\linewidth]{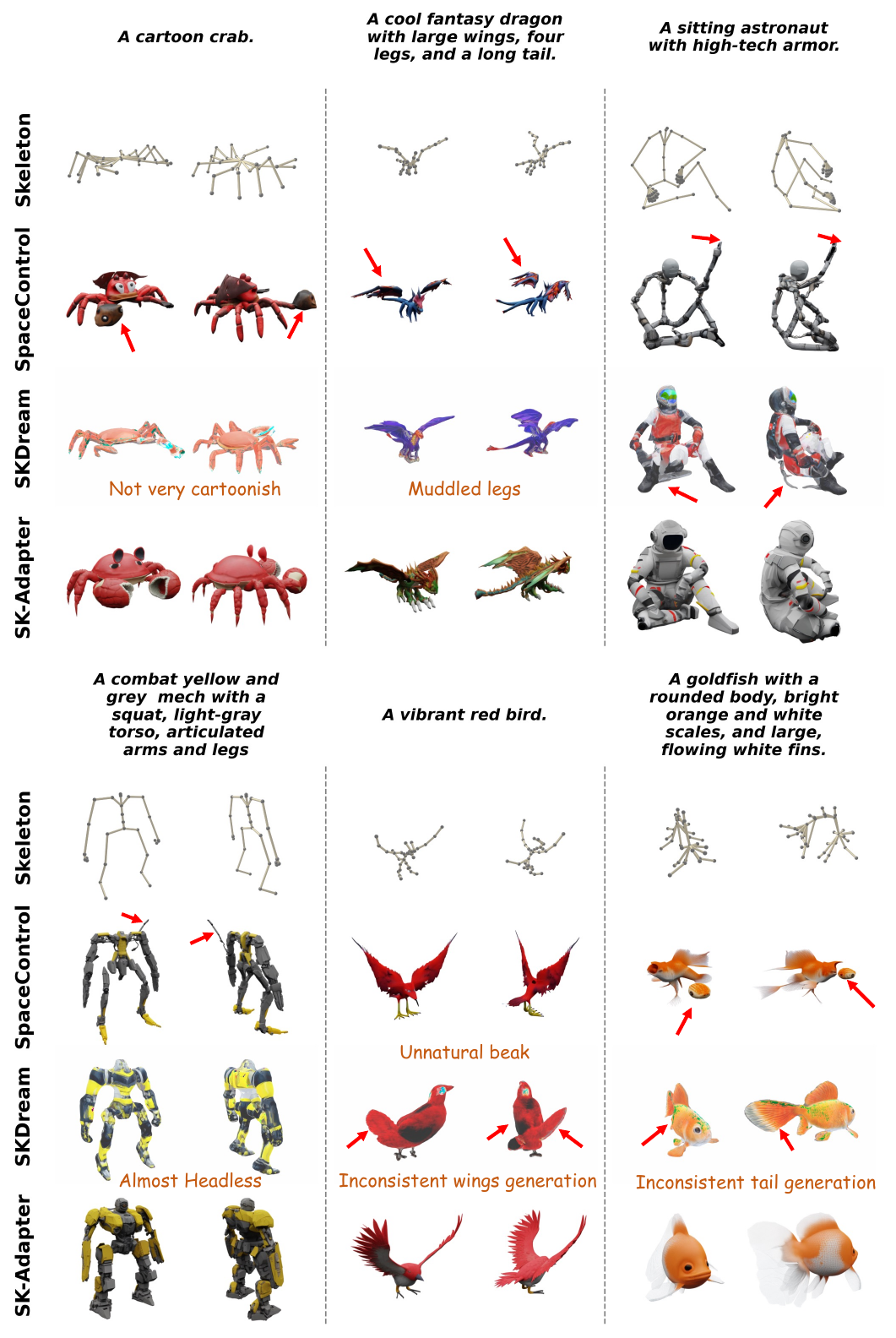}
    \caption{Qualitative comparison of our SK-Adapter with the baselines. For the baseline generations, arrows indicate structural inconsistency with the given skeleton. Short captions indicate one of the apparent problems in their generated results.}
    \label{fig:comparison}
\end{figure}

 \begin{figure}[t]
    \centering
    \includegraphics[width=1.0\linewidth]{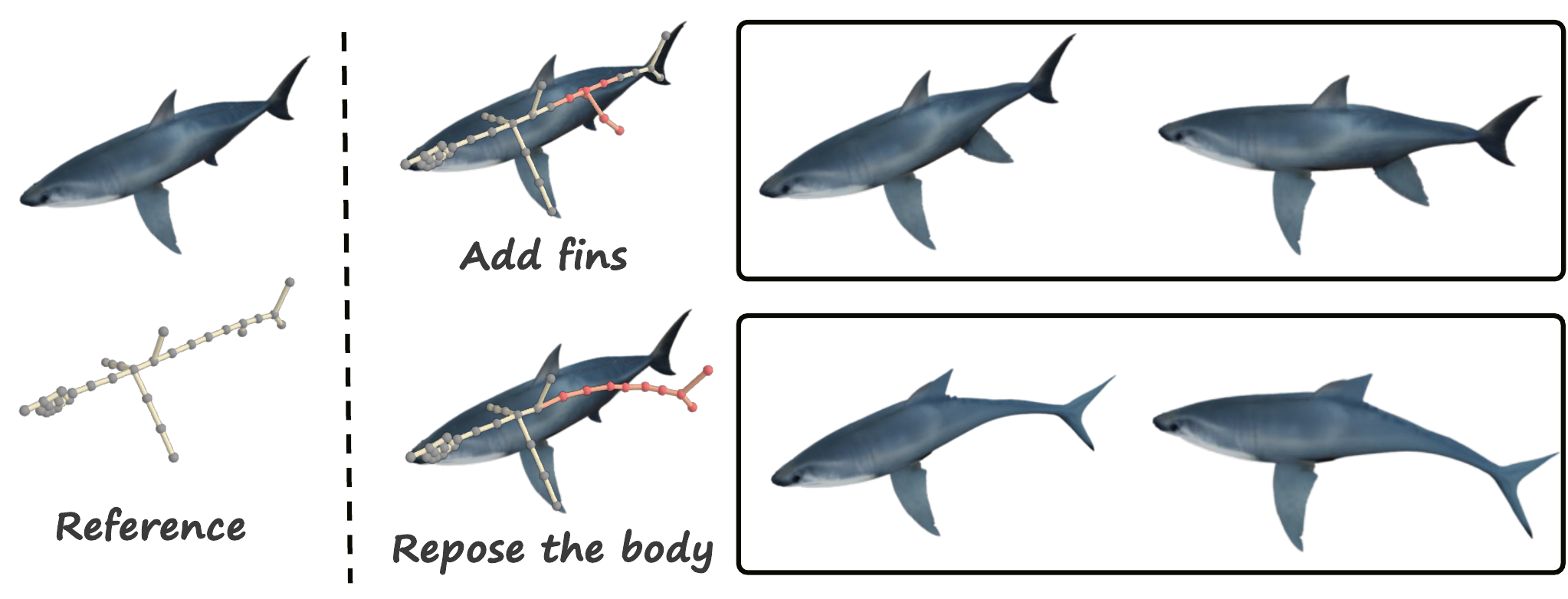} 
    \caption{SK-Adapter enables flexible editing with skeleton, including addition and re-posing.}
    \label{fig:edit}
\end{figure}

\subsection{Quantitative Results}

The quantitative comparisons on TMS-eval are summarized in Tables~\ref{tab:alignment_results} and~\ref{tab:visual_results}. 

\textbf{Structural and Textual Alignment.~} Evidenced by Table~\ref{tab:alignment_results}, SK-Adapter achieves state-of-the-art performance in both structural fidelity and semantic consistency. Our method drastically reduces the overall ReRigging Score to 0.2228, significantly outperforming the lifting-based baseline SKDream with 0.2818 and the tuning-free spatial guidance method SpaceControl with 0.2740. This substantial margin confirms that injecting skeletal tokens within the native 3D space effectively mitigates the spatial ambiguity inherent in 2D projection strategies, resulting in assets that faithfully 
conform to the user-defined topological constraints without overdoing it. Furthermore, SK-Adapter achieves the highest overall CLIP Score, demonstrating that our lightweight adapter maintains robust text-to-3D cross-modal alignment.

\textbf{Visual Fidelity.~} Beyond structural precision, our framework maintains superior visual quality, as detailed in Table~\ref{tab:visual_results}. SK-Adapter achieves the highest overall PickScore of 21.01, indicating a strong human preference for our generated geometries. Also, our method significantly lowers the KD-DINO metric to 0.7778, compared to 1.3809 for SKDream and 1.7821 for SpaceControl. These results strongly indicate that our strategy of freezing the backbone effectively preserves the powerful generative priors of the foundation model, synthesizing high-frequency details and coherent textures without suffering from the geometric artifacts or texture degradation typically observed in multi-stage reconstruction pipelines.

\textbf{Running Time.~} Under the same hardware conditions, our SK-Adapter as well as SpaceControl can generate a 3D asset in less than $15$ seconds while SKDream requires around $40$ seconds to generate one sample on average.

\subsection{Qualitative Results}
Visual comparisons are presented in Fig. \ref{fig:comparison}. Our method consistently outperforms baseline approaches by demonstrating significantly better skeleton interpretation, finer structural alignment, and stricter adherence to text prompts. Furthermore, it yields higher visual fidelity and more detailed geometries. While our approach excels at generating coherent and intricate structures, alternative methods suffer from notable quality degradation. SKDream exhibits structural distortions and vague or missing texture details caused by the multi-view inconsistencies inherent in its underlying 2D generative models; SpaceControl produces assets that are overly constrained by the skeleton, often leading to fragmented or broken topologies.
\subsection{Application}
\subsubsection{Editing.}
Fig.~\ref{fig:edit} illustrates sample editing sequences of 3D assets, involving addition and re-posing (replacement). Given an edited skeleton with the identified bounding box, our method enables skeleton-based flexible editing on existing 3D assets, where the generated mesh of the locally-edited skeleton blends seamlessly with the original mesh, preserving the underlying structural and textural coherence.
\subsubsection{Animation.} 
Our skeleton-conditioned generated 3D assets can be seamlessly integrated into standard animation pipelines. Our approach outputs static 3D assets that can be rigged and driven by skeletal motions. Specifically, we use the off-the-shelf skinning model from Anymate\cite{deng2025anymatedatasetbaselineslearning} to compute skinning weights using the input skeleton and our generated 3D assets. Given the skinning weights, we apply Linear Blend Skinning (LBS) to deform the asset according to the input skeleton sequences.

\begin{figure}
    \centering
    \includegraphics[width=1.0\linewidth]{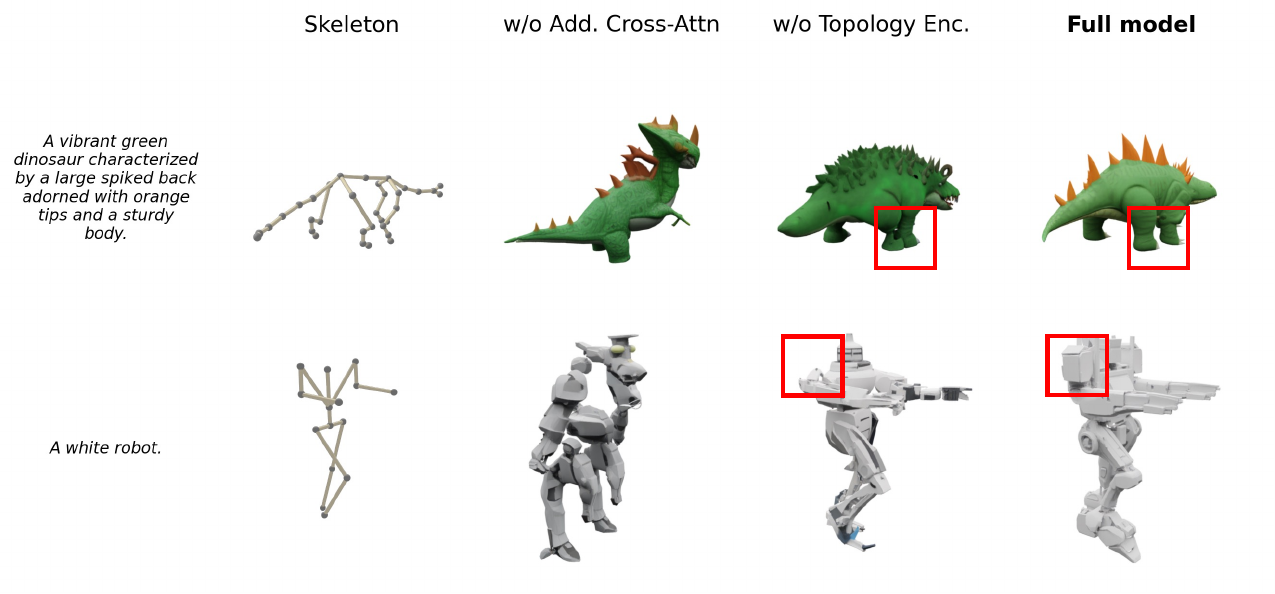}
    \caption{Comparison of different architecture designs. Without Cross-Attention, the model tends to collapse given complex skeleton conditions. In contrast, the topology-aware encoder helps the model better capture skeleton structures, as indicated by the bounding boxes.}
    \label{fig:compare_ablation}
\end{figure}

\subsection{Ablation Study}
To validate the effectiveness of our key architectural components, we conduct an ablation study by evaluating different variants of SK-Adapter. Each variant is trained on an 8k subset of Objaverse-TMS for 200 epochs.
Specifically, we implement the ablation study on the following components:

\textbf{Skeletal Cross Attention.} In this variant, we remove the dedicated skeletal cross-attention layers. Instead, we directly concatenate the encoded skeleton features with the text features, thus forcing both modalities to share a single cross-attention layer.
In doing so, we evaluate the necessity of a dedicated cross-attention mechanism for injecting structural guidance without entangling it with textual semantics.

\textbf{Topology-Aware Encoding.} In this variant configuration, the topological graph input and GRPE encoder are removed. Consequently, the network is conditioned solely on the positional embeddings of the spatial joint coordinates. This tests the importance of explicitly modeling hierarchical connectivity and topological constraints for accurate 3D articulation.

Tables~\ref{tab:ablation_alignment} and~\ref{tab:ablation_visual} tabulate the quantitative results tested on TMS-eval. 
As shown in Table~\ref{tab:ablation_alignment}, the Full Model achieves the most robust structural control, yielding the best ReRigging Score of 0.2355 overall. Removing the dedicated skeletal cross-attention causes a severe degradation in structural alignment, effectively doubling the ReRigging Score to 0.5049. Similarly, discarding the topological encoding negatively impacts the model's ability to accurately articulate the 3D generation.

Crucially, Tables~\ref{tab:ablation_alignment} and~\ref{tab:ablation_visual} demonstrate that the Full Model's significant improvements in structural control do not come at the expense of generation quality. Although the variant without topology encoding shows very marginal gains in CLIP Score and KD-DINO, the Full Model achieves the highest PickScore across all categories. The Full Model therefore offers the optimal architectural balance between fine-grained skeleton controllability and high-quality 3D generation.

Qualitative comparisons are shown in Fig.~\ref{fig:compare_ablation}. Without Cross-Attention, the model tends to collapse given complex skeleton conditions. Although the model without a topological encoder can generate 3D assets that follow the skeleton joints, the topology-aware encoder helps it better capture skeleton structures.

\begin{table*}[t]
\centering
\caption{\textbf{Ablation Study (Structural and Textual Alignment).} Impact of individual architectural components on structural control and textual alignment across overall, humanoid, and animal categories. $\downarrow$ indicates lower is better, $\uparrow$ indicates higher is better. \textbf{Bold} indicates the best performance, \underline{underline} indicates the second best.}
\label{tab:ablation_alignment}
\begin{tabular}{l|ccc|ccc}
\toprule
 & \multicolumn{3}{c|}{\textbf{ReRigging Score} $\downarrow$} & \multicolumn{3}{c}{\textbf{CLIP Score} $\uparrow$} \\
\cmidrule(lr){2-4} \cmidrule(lr){5-7}
\textbf{Variant} & Overall & Humanoid & Animal & Overall & Humanoid & Animal \\
\midrule
w/o Add. Cross-Attn & 0.5049 & 0.4230 & 0.5353 & 24.71 & 25.16 & 25.27 \\
w/o Topology Enc.   & \underline{0.2527} & \underline{0.1885} & \underline{0.2747} & \textbf{26.14} & \textbf{26.42} & \underline{26.87} \\
\midrule
\textbf{Full Model} & \textbf{0.2355} & \textbf{0.1832} & \textbf{0.2531} & \underline{26.11} & \underline{26.16} & \textbf{26.99} \\
\bottomrule
\end{tabular}%

\caption{\textbf{Ablation Study (Visual Fidelity).} Impact of individual architectural components on visual fidelity across overall, humanoid, and animal categories. $\downarrow$ indicates lower is better, $\uparrow$ indicates higher is better. \textbf{Bold} indicates the best performance, \underline{underline} indicates the second best.}
\label{tab:ablation_visual}
\begin{tabular}{l|ccc|ccc}
\toprule
 & \multicolumn{3}{c|}{\textbf{PickScore} $\uparrow$} & \multicolumn{3}{c}{\textbf{KD-DINO} $\downarrow$} \\
\cmidrule(lr){2-4} \cmidrule(lr){5-7}
\textbf{Variant} & Overall & Humanoid & Animal & Overall & Humanoid & Animal \\
\midrule
w/o Add. Cross-Attn & 20.54 & 20.39 & 20.86 & 1.0914 & 1.2346 & 1.9521 \\
w/o Topology Enc.   & \underline{20.90} & \underline{20.61} & \underline{21.29} & \textbf{0.7865} & \textbf{1.1539} & \textbf{1.3378} \\
\midrule
\textbf{Full Model} & \textbf{20.94} & \textbf{20.61} & \textbf{21.36} & \underline{0.8616} & \underline{1.1655} & \underline{1.3812} \\
\bottomrule
\end{tabular}%

\end{table*}

\section{Conclusion}
\label{sec:conclusion}
In this paper, we propose SK-Adapter, a lightweight framework that achieves precise skeleton-guided structural control for native 3D generation. By encoding 3D skeletons as sparse, topology-aware spatial tokens and seamlessly injecting them into a frozen 3D flow transformer, our method ensures faithful structural alignment while fully preserving the powerful generative priors of the foundation model. These capabilities directly facilitate the creation of production-ready, articulable 3D assets. Consequently, by treating skeletons as control signals, SK-Adapter bridges the gap between abstract semantic synthesis and explicit topological articulation, paving the way toward more interpretable and controllable 3D generation.

%
%
\bibliographystyle{splncs04}
\bibliography{main}

---- Appendix ----
\newpage
\setcounter{page}{1}
\onecolumn
\appendix
\section*{Appendix}
\section{Trellis}
\textbf{Rectified Flow Matching.} 
TRELLIS is built upon the Rectified Flow framework, which learns a generative model by transforming a noise distribution into a data distribution through a deterministic straight-line path. 

Specifically, let $x_0 \sim p_{\text{data}}$ be a sample from the target 3D data distribution and $\epsilon \sim \mathcal{N}(0, \mathbf{I})$ be Gaussian noise. Rectified Flow defines a linear interpolation $x_t$ between the data and noise for $t \in [0, 1]$:
$$x_t = (1 - t)x_0 + t\epsilon$$
The target velocity field that drives the transformation from noise to data is the time-derivative of $x_t$:
$$v(x_t, t) = \frac{dx_t}{dt} = \epsilon - x_0$$
A neural network $v_\theta(x_t, t)$ is trained to approximate this velocity field by minimizing the Conditional Flow Matching (CFM) objective:
$$\mathcal{L}_{\text{CFM}}(\theta) = \mathbb{E}_{t, x_0, \epsilon} \left\| v_\theta(x_t, t) - (\epsilon - x_0) \right\|^2_2$$
During inference, given a noise $\epsilon$ and a condition $c$, the model generates a sample by solving an Ordinary Differential Equation (ODE) from $t=1$ to $t=0$ along the predicted velocity field.

TRELLIS addresses the sparsity and complexity of 3D data by generating a Structured LATent (SLAT) representation. SLAT defines a set of local latents $\{(z_i, p_i)\}_{i=1}^L$ on a 3D grid, where $p_i$ is the index of an active voxel intersecting the object surface, and $z_i \in \mathbb{R}^C$ is the feature vector capturing local geometry and appearance. TRELLIS generates SLAT via a specialized two-stage pipeline:

\textbf{Sparse Structure Generation.} 
A noisy latent variable $\mathbf{z}_1 \in \mathbb{R}^{D \times D \times D \times C_S}$ is sampled from a standard Gaussian distribution $\mathcal{N}(\mathbf{0}, \mathbf{I})$ and iteratively denoised to $\mathbf{z}_0$ through a rectified flow model ($G_S$). This process is conditioned on global prompts, such as text features from CLIP or visual features from DINOv2. The clean latent $\mathbf{z}_0$ is then decoded by a pre-trained decoder $\mathcal{D}$ into a binary occupancy grid $\mathbf{x} \in \{0, 1\}^{V \times V \times V}$, which explicitly defines the coarse spatial structure of the 3D asset. It is important to note that decoder $\mathcal{D}$ has a corresponding VAE encoder $\mathcal{E}$ which was not used in inference, but is very crucial for our fine-tuning.

\textbf{Structured Latents Generation.}
Given the spatial structure $\mathbf{x}$ from the first stage, the Sparse Flow Transformer ($G_L$) synthesizes the local feature vectors for each active voxel. For each coordinate $p_i \in \mathbf{x}$, a local noisy latent $h_{1,i} \in \mathbb{R}^C$ is sampled. $G_L$ then performs a second rectified flow process to denoise these sparse latents into $h_{0,i}$, conditioned on both the global prompts and the generated structure $\mathcal{P}$. This step ensures that the synthesized features are spatially coherent with the predicted layout. Finally, the complete set of structured latents $\{(h_{0,i}, p_i)\}$ is mapped to specific 3D representations, such as 3D Gaussians or Meshes, via a suite of pre-trained, representation-specific decoders.

\section{Topological Encoding Details}
\label{appendix:topology_codebook}

In this appendix, we provide a detailed specification of the two graph-structural codebooks used in our Graph Relative Positional Encoding (GRPE): the \textit{topological distance} codebook and the \textit{edge relation} codebook. Given a skeleton represented as a rooted tree with parent array $\mathbf{p} \in \mathbb{Z}^N$ (where $p_i$ denotes the parent index of joint $i$, and $p_r = -1$ for the root $r$), we construct two pairwise matrices $\mathcal{D}_S \in \mathbb{Z}^{N \times N}$ and $\mathcal{R}_S \in \mathbb{Z}^{N \times N}$.

\subsection{Topological Distance Codebook $E^D$}

The topological distance $D_{ij}$ between joints $i$ and $j$ is defined as the length of the shortest path connecting them in the skeleton tree. To keep the embedding table compact and prevent overfitting to rare long-range pairs, we clip all distances to a maximum path length $d_{\text{max}}$:
\begin{equation}
    D_{ij} = \min\bigl(\text{dist}_{\mathcal{G}}(i, j),\; d_{\text{max}}\bigr),
\end{equation}
where $\text{dist}_{\mathcal{G}}(i,j)$ counts the number of edges on the unique tree path from $i$ to $j$. In our experiments, we set $d_{\text{max}} = 5$, yielding $d_{\text{max}} + 1 = 6$ discrete distance levels as enumerated in Table~\ref{tab:topo_dist}.

\begin{table}[h]
\centering
\caption{Topological distance codebook entries.}
\label{tab:topo_dist}
\begin{tabular}{cl}
\toprule
\textbf{Index} & \textbf{Semantics} \\
\midrule
0 & Self-loop ($i = j$) \\
1 & Direct parent or child (1-hop neighbor) \\
2 & Grandparent or grandchild (2-hop) \\
3 & 3-hop ancestor/descendant \\
4 & 4-hop ancestor/descendant \\
5 & $\geq 5$-hop or disconnected (clipped) \\
\bottomrule
\end{tabular}
\end{table}

We learn three independent embedding matrices for the distance codebook:
\begin{itemize}
    \item $E_q^D \in \mathbb{R}^{(d_{\text{max}}+1) \times F}$: query-side distance embedding,
    \item $E_k^D \in \mathbb{R}^{(d_{\text{max}}+1) \times F}$: key-side distance embedding,
    \item $E_v^D \in \mathbb{R}^{(d_{\text{max}}+1) \times F}$: value-side distance embedding.
\end{itemize}
Each embedding row is looked up by the clipped distance index $D_{ij}$ and injected into the corresponding attention component as described in the main text.

\subsection{Edge Relation Codebook $E^R$}

While the distance codebook captures how \textit{far apart} two joints are, the edge relation codebook encodes the \textit{semantic role} of their relationship. We define six mutually exclusive relation types, summarized in Table~\ref{tab:edge_rel}.

\begin{table}[h]
\centering
\caption{Edge relation codebook entries. For each ordered pair $(i, j)$, exactly one relation type is assigned.}
\label{tab:edge_rel}
\begin{tabular}{cll}
\toprule
\textbf{Index} & \textbf{Type} & \textbf{Definition} \\
\midrule
0 & \textsc{Self} & $i = j$ and $i$ has at least one child \\
1 & \textsc{Parent} & $j = p_i$ (joint $j$ is the parent of $i$) \\
2 & \textsc{Child} & $p_j = i$ (joint $j$ is a child of $i$) \\
3 & \textsc{Sibling} & $p_i = p_j$ and $i \neq j$ (share the same parent) \\
4 & \textsc{Distant} & None of the above (no direct kinematic relation) \\
5 & \textsc{End-Effector} & $i = j$ and $i$ has no children (leaf node) \\
\bottomrule
\end{tabular}
\end{table}

The relation assignment follows the precedence order listed above; specifically, the diagonal entry $R_{ii}$ is set to \textsc{End-Effector} (5) if joint $i$ is a leaf node in the skeleton tree, and to \textsc{Self} (0) otherwise. The distinction between \textsc{Self} and \textsc{End-Effector} is motivated by the special kinematic role of leaf joints: end effectors (\eg, fingertips, toes, head) serve as the primary targets in inverse kinematics and are the points most directly constrained by user-specified poses. Providing a dedicated embedding allows the model to learn this structural prior.

The off-diagonal entries encode the \textit{directed} kinematic relationship. Note that the relation matrix is asymmetric: if $R_{ij} = \textsc{Parent}$, then $R_{ji} = \textsc{Child}$. The \textsc{Sibling} relation is symmetric and captures branching points in the skeleton hierarchy (\eg, left and right shoulders sharing the spine as their parent).

Analogous to the distance codebook, we learn three embedding matrices:
\begin{itemize}
    \item $E_q^R \in \mathbb{R}^{6 \times F}$: query-side relation embedding,
    \item $E_k^R \in \mathbb{R}^{6 \times F}$: key-side relation embedding,
    \item $E_v^R \in \mathbb{R}^{6 \times F}$: value-side relation embedding.
\end{itemize}

\subsection{Summary}

In total, the topology codebook comprises six learnable embedding matrices (three for distance, three for relations), containing $(d_{\text{max}} + 1 + 6) \times 3 \times F = 36F$ parameters. For $F = 1024$, this amounts to only $\sim$37K additional parameters---a negligible overhead compared to the backbone. Despite this compactness, the codebook provides the attention mechanism with rich structural inductive biases: the distance codebook captures the notion of \textit{proximity} along the kinematic chain, while the relation codebook encodes \textit{directionality} (parent vs.\ child) and \textit{branching structure} (siblings, end effectors) that are essential for understanding articulated body topology.
\newpage

\section{SK-Adapter Configuration}
\label{appendix:params}

\begin{figure}
    \centering
    \includegraphics[width=0.7\linewidth]{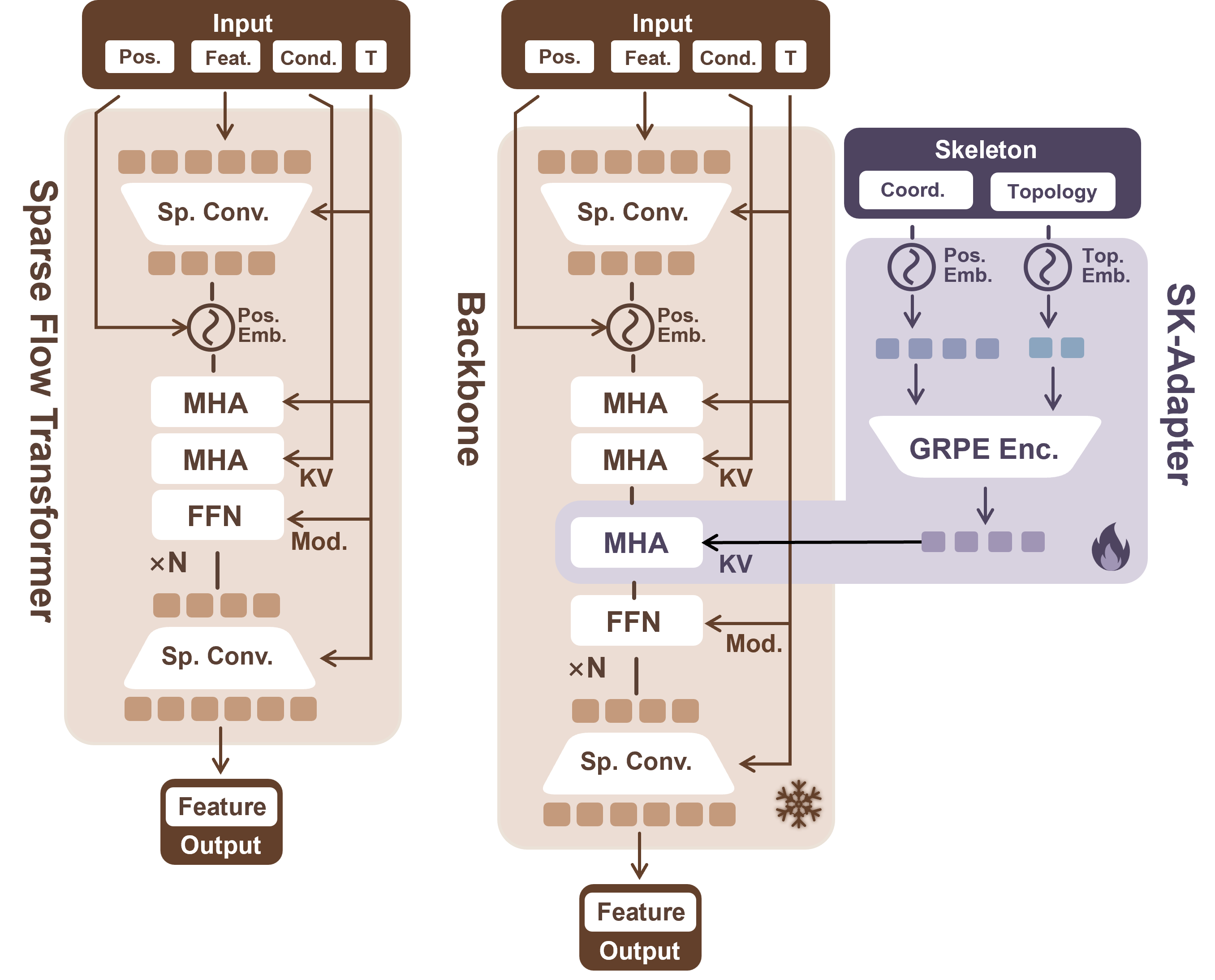}
    \caption{Comparison between Trellis's Structure Flow Transformer and our SK-Adapter. Backbone and SK-Adapter is connected via an additional cross-attention layer in each DiT block.}
    \vspace{-4mm}
    \label{fig:network}
\end{figure}

Table~\ref{tab:param_budget} summarizes the trainable parameter count of SK-Adapter, with $F{=}1024$ and the TRELLIS-text-large backbone ($L{=}24$ blocks). The architecture comparison is shown in Fig. \ref{fig:network}.

\begin{table}[h]
\centering
\caption{Trainable parameter breakdown of SK-Adapter.}
\label{tab:param_budget}
\begin{tabular}{lcrr}
\toprule
\textbf{Layer} & \textbf{Count} & \textbf{Params / Unit} & \textbf{Total} \\
\midrule
Topology-Aware Attention & $\times 2$ & 4,235,264 & 8,470,528 \\
Feed-Forward Network & $\times 2$ & 8,393,728 & 16,787,456 \\
LayerNorms (Encoder) & $\times 4$ & 2,048 & 8,192 \\
\midrule
Skeletal Cross-Attention & $\times 24$ & 4,198,400 & 100,761,600 \\
Zero-Initialized Linear & $\times 24$ & 1,049,600 & 25,190,400 \\
LayerNorms (Adapter) & $\times 48$ & 2,048 & 98,304 \\
\midrule
\textbf{Total Trainable} & & & \textbf{151,316,480} \\
\bottomrule
\end{tabular}
\end{table}

\noindent The topology codebook embeddings (6 tables $\times$ 2 layers $= 73{,}728$ params) are included in the Topology-Aware Attention row. The total adapter overhead is ${\sim}151$M trainable parameters.  

\newpage
\section{More Visualizations}
\begin{figure}
    \centering
    \includegraphics[width=0.93\linewidth]{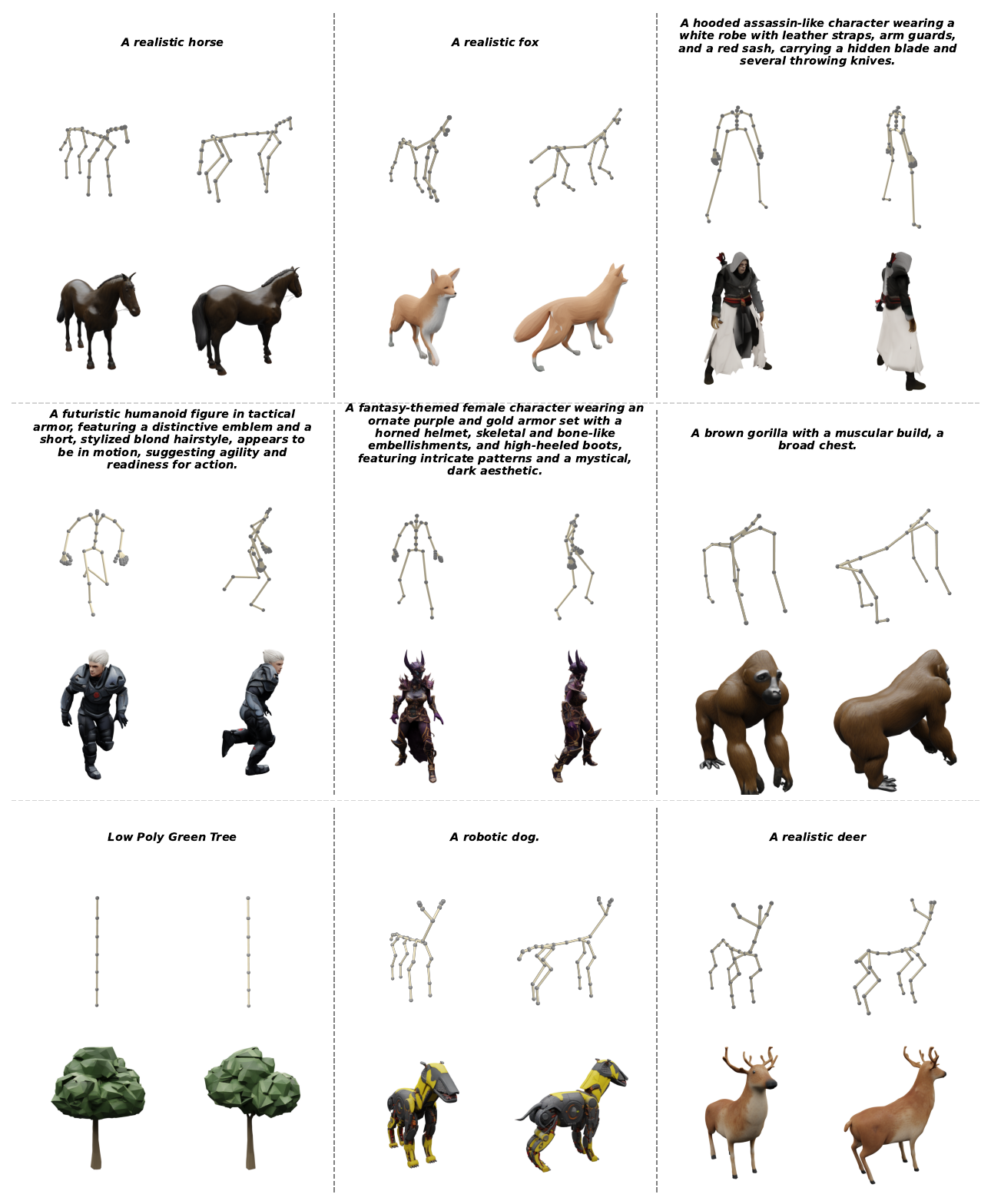}
    \caption{Additional qualitative results generated with SK-Adapter.}
    \label{fig:moresamples}
\end{figure}
\newpage
\section{Visualizations for Editing}
\begin{figure}
    \centering
    \includegraphics[width=0.93\linewidth]{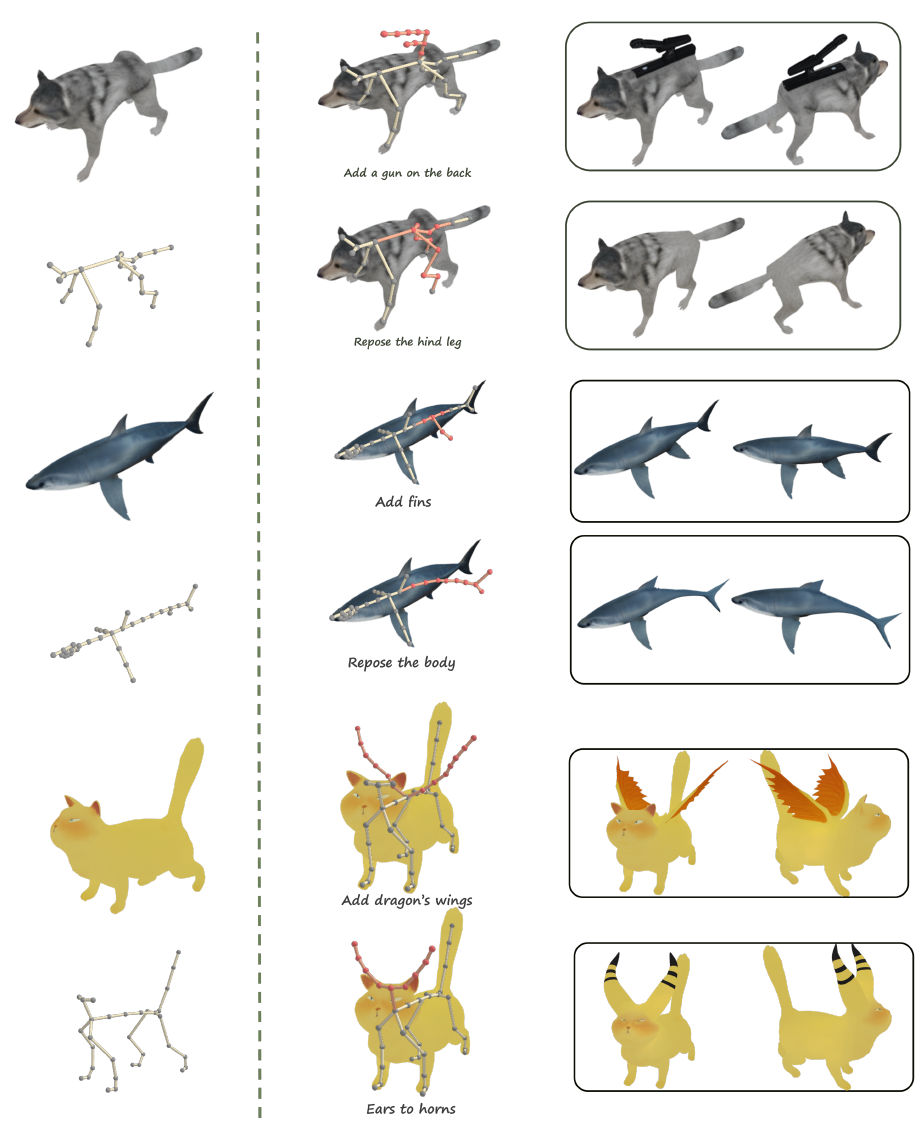}
    \caption{Visualization of editing examples}
    \label{fig:editing_vis}
\end{figure}

\newpage
\section{Limitation}
\begin{figure}
    \centering
    \includegraphics[width=1.0\linewidth]{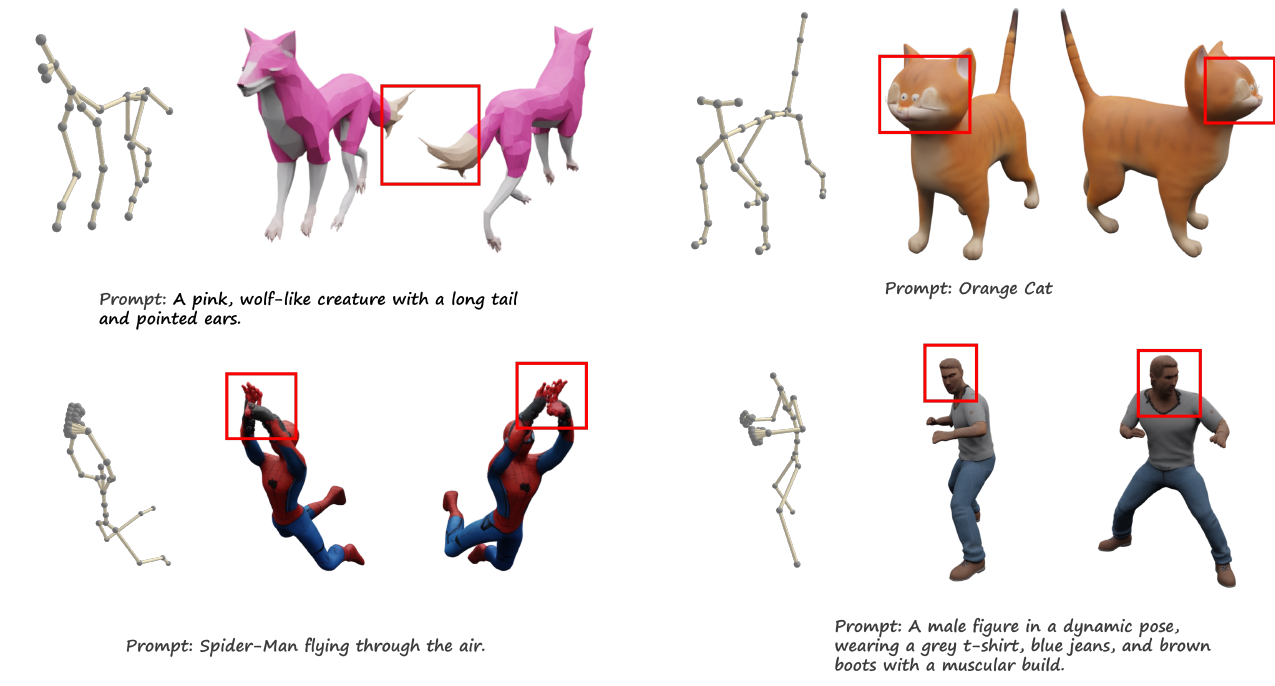}
    \caption{Some typical limitations of base model and SK-Adapter}
    \label{fig:limitation}
\end{figure}
\label{sec:limitation}
While our proposed SK-Adapter demonstrates strong capabilities in generating native 3D assets that strictly conform to user-specified skeletal structures, the overall generation quality and fidelity remain intrinsically bounded by the capabilities of the pre-trained 3D base model. As illustrated in the failure cases in Fig.~\ref{fig:limitation}, several typical limitations can be observed:

It occasionally struggles with fine-grained geometric and textural details. For instance, the generated tail of the wolf-like creature appears rough and lacks natural structural coherence. Also, The base model's limitations are evident in region-sensitive areas such as faces. As shown in the generations of the orange cat and the human figure, the facial features are often distorted with wrong structure. Meanwhile, the semantic alignment between the text prompt and the generated 3D asset is fundamentally limited by the underlying 3D base model and the text encoder. Consequently, some complex or highly specific attributes described in the prompt may not be perfectly translated into the final 3D geometry and texture.
 
When the conditioning skeleton become too complex, the structural guidance can become ambiguous. A notable example is the hand region of the Spider-Man figure, where the intricate and intersecting topology of the fingers causes the model to produce messy and poorly resolved local geometries.

Future work could address these limitations by integrating more advanced, higher-resolution 3D foundation models, utilizing more powerful text encoder for enhanced text-to-3D alignment and training SK-Adapter on larger scale.

\end{document}